\documentclass[lettersize,journal]{IEEEtran}
\usepackage{amsmath,amsfonts,amssymb}
\usepackage[ruled]{algorithm2e}
\usepackage{array}
\usepackage{bbm}
\usepackage[caption=false,font=normalsize,labelfont=sf,textfont=sf]{subfig}
\usepackage{textcomp}
\usepackage{stfloats}
\usepackage{url}
\usepackage{verbatim}
\usepackage{graphicx}
\usepackage{booktabs}
\usepackage{multirow} 
\usepackage{cite}
\usepackage{hyperref} 
\usepackage{color}

\makeatletter

\newcommand{\Rmnum}[1]{\expandafter\@slowromancap\romannumeral #1@}
\makeatother

\begin{document}

\title{Cooperative Decision-Making for CAVs at Unsignalized Intersections: A MARL Approach with Attention and Hierarchical Game Priors}

\author{Jiaqi Liu,~\IEEEmembership{Student Member,~IEEE,} Peng Hang,~\IEEEmembership{Member,~IEEE,} Xiaoxiang Na,
\\Chao Huang,~\IEEEmembership{Senior Member,~IEEE,} and Jian Sun
\thanks{This work was supported in part by the National Key R\&D Program of China (2023YFB4301900), the Shanghai Scientific Innovation Foundation (No.23DZ1203400), the National Natural Science Foundation of China (52302502), the State Key Laboratory of Intelligent Green Vehicle and Mobility under Project No. KFZ2408, and the Fundamental Research Funds for the Central Universities.}
\thanks{Jiaqi Liu, Peng Hang, and Jian Sun are with the College of Transportation and Key Laboratory of Road and Traffic Engineering,
Ministry of Education, Tongji University, Shanghai 201804, China. (e-mail: \{liujiaqi13, hangpeng, sunjian\}@tongji.edu.cn)}
\thanks{Xiaoxiang Na is with the Department of Engineering, University of Cambridge, Cambridge CB2 1PZ, United Kingdom (e-mail: xnhn2@cam.ac.uk )}
\thanks{Chao Huang is with Department of Industrial and System Engineering, The Hong Kong Polytechnical University, Hong Kong 999077. (e-mail:  hchao.huang@polyu.edu.hk) }
\thanks{Corresponding author: Peng Hang}
}

\markboth{}%
{Shell \MakeLowercase{\textit{et al.}}: A Sample Article Using IEEEtran.cls for IEEE Journals}

\maketitle

\begin{abstract}
The development of autonomous vehicles has shown great potential to enhance the efficiency and safety of transportation systems. However, the decision-making issue in complex human-machine mixed traffic scenarios, such as unsignalized intersections, remains a challenge for autonomous vehicles. While reinforcement learning (RL) has been used to solve complex decision-making problems, existing RL methods still have limitations in dealing with cooperative decision-making of multiple connected autonomous vehicles (CAVs), ensuring safety during exploration, and simulating realistic human driver behaviors. 
In this paper, a novel and efficient algorithm, Multi-Agent Game-prior Attention Deep Deterministic Policy Gradient (MA-GA-DDPG), is proposed to address these limitations. Our proposed algorithm formulates the decision-making problem of CAVs at unsignalized intersections as a decentralized multi-agent reinforcement learning problem and incorporates an attention mechanism to capture interaction dependencies between ego CAV and other agents. The attention weights between the ego vehicle and other agents are then used to screen interaction objects and obtain prior hierarchical game relations, based on which a safety inspector module is designed to improve the traffic safety. Furthermore, both simulation and hardware-in-the-loop experiments were conducted, demonstrating that our method outperforms other baseline approaches in terms of driving safety, efficiency, and comfort.
\end{abstract}

\begin{IEEEkeywords}
Multi-agent reinforcement learning, connected autonomous vehicles, decision-making, attention mechanism, unsignalized intersections
\end{IEEEkeywords}

\section{Introduction}
\IEEEPARstart{A}{utonomous} vehicles (AVs) have undergone remarkable advances in recent years, holding great potential for enhancing transportation efficiency and safety \cite{aradi2020survey,lu2022autonomous}.
Nonetheless, decision-making in complex human-machine mixed traffic scenarios, particularly at unsignalized intersections, remains a considerable challenge for both AVs and human drivers \cite{rahmati2021helping}. Game-based and optimization-based approaches have been proposed to address this issue \cite{yuan2021deep,hang2022decision,pan2022convex}. However, these methods prove impractical when handling scenarios involving multiple agents and complex interaction behaviors \cite{pan2022convex}. Reinforcement learning (RL) holds the potential to overcome these limitations, leveraging data-driven methods with its robust learning and efficient reasoning capabilities \cite{wu2021flow,zhang2022safe,liu2023mtd}.

Current RL approaches for unsignalized intersection decision-making problems encounter several challenges. Firstly, most studies consider a single AV at the intersection, modeling the problem as a single-agent RL problem, whereas cooperative decision-making of multiple connected autonomous vehicles (CAVs) is a more challenging and less explored problem. Besides, common RL methods rely on exploration to teach CAVs how to decide and act, which may compromise the learning efficiency and policy safety \cite{aradi2020survey}. Safety is the most critical factor when designing the decision-making algorithm. Moreover, the simulation environments' simplification of human drivers' behaviors may lead to performance gaps between simulation and real-world scenarios.

To cope with these challenges, we propose a novel and efficient algorithm, Multi-Agent Game-prior Attention Deep Deterministic Policy Gradient (MA-GA-DDPG), which formulates the decision-making problem of CAVs at unsignalized, human-machine mixed intersections as a decentralized multi-agent RL problem. In MA-GA-DDPG, each CAV at the intersection is modeled as an agent, enabling it to explore the environment, communicate, and cooperate with other agents. We use Multi-Agent Deep Deterministic Policy Gradient (MADDPG) as the baseline algorithm, where all agents adopt a strategy of centralized training and distributed execution (CTDE). To capture the interaction dependencies between the ego CAV and other agents, we incorporate an attention mechanism \cite{vaswani2017attention}. The attention weights between the ego vehicle and other agents are used to screen interaction objects and obtain prior hierarchical game relations. We then develop a safety inspector module to predict and detect potential conflicts with other agents during CAV exploration and make corrections in real-time to improve the algorithm's learning efficiency. We also take into account the heterogeneity of human drivers in traffic environments and carry out extensive experiments using both simulation and hardware-in-loop setups, during which domain controllers are utilized for decision-making and planning. The results demonstrate that our approach excels in terms of learning efficiency, driving safety, as well as overall efficiency and comfort.

The contributions of this paper are summarized as follows:
\begin{itemize}
    \item A novel and efficient algorithm MA-GA-DDPG is proposed to realize cooperative decision-making of CAVs in complex human-machine mixed traffic scenarios. The heterogeneity of drivers is considered to simulate the real traffic environment and train the algorithm.
    \item The multi-head attention mechanism is leveraged to capture the complex interactions among CAVs and other vehicles in the mixed driving environment. An interaction object filter based on the attention weights is designed to facilitate the identification of the most relevant agents for interaction.
    \item A hierarchical game framework is utilized to incorporate the traffic priority priors into the interaction process. This framework enables the modeling of the priority information and its subsequent transmission to the safety inspector module, which effectively supervises and adjusts the actions of CAVs to minimize the risk of collisions.
\end{itemize}

The rest of the paper is organized as follows: Section \ref{section:2} summarizes the recent related works. The decision-making problem at the intersection is formulated in section \ref{section:3}. In section \ref{section:4}, the MA-GA-DDPG model is described. In section \ref{section:5}, the simulation environment and comprehensive experiments are introduced and the results are analyzed. Finally, this paper is concluded in section \ref{section:6}.

\section{Related Works}
\label{section:2}
\subsection{Decision-Making of CAVs at Intersections}
Navigating complex, high-density mixed driving conditions presents a formidable challenge for CAVs, with intersections ranking among the most intricate and demanding scenarios for interaction in autonomous driving \cite{leurent2019social,wei2021autonomous}. Developing an efficient cooperative decision-making algorithm tailored to intersection scenarios holds paramount importance in enhancing intersection traffic's efficiency and safety \cite{guillen2020raim}.

Current research in decision-making models for autonomous driving at unsignalized intersections encompasses several approaches:

\begin{itemize}
    \item Game-theoretical models, such as Level-k games\cite{yuan2021deep}, follower-leader games\cite{li2020game}, etc. These studies treat each CAV as a rational decision-maker and simulate the reactions and actions of human drivers under rational conditions\cite{9781400}. However, this purely rational situation cannot fully simulate the real world. Moreover, the efficiency of large-scale game calculations and the scalability of game frameworks are also challenging issues.

    \item Rule-based methods, such as first-come-first-serve (FCFS)\cite{dresner2004multiagent}, Buffer Coordination\cite{lin2017autonomous} etc. These methods are easy to implement and logically clear, but as the traffic demand increases, the efficiency of these methods is poor.
    
    \item Optimization-based methods, such as convex optimization methods\cite{pan2022convex,kamal2014vehicle}, model predictive control(MPC)\cite{schildbach2016collision} etc. The advantage of this method is that it can be solved accurately, with good interpretability and controllability, but the solution efficiency of too large-scale problems often cannot meet the requirements of real-time applications\cite{kamal2014vehicle}.
    
    \item Learning-based models, encompassing Neural Networks (NN) \cite{hecker2018end} and Reinforcement Learning (RL) \cite{xue2022multi, daisocially, zhang2022safe}, effectively capture interaction dynamics with potent learning and efficient reasoning capabilities \cite{wang2021highway,chen2021deep}. However, addressing challenges related to interpretability, convergence, and generalization remains essential \cite{oroojlooy2022review}.
    
\end{itemize}

\subsection{Multi-Agent Reinforcement Learning and Attention Mechanism}
Multi-Agent Reinforcement Learning (MARL) is an emerging research field that focuses on the optimization problem of multiple autonomous intelligent agents making sequential decisions in an environment. In recent years, MARL has been utilized to solve plenty of multi-agent problems, such as traffic control\cite{zhang2019cityflow}, decision-making of autonomous driving\cite{wang2021highway,chen2021deep,toghi2022social}, games\cite{bard2020hanabi},  etc. 

Some works have modeled the traffic system with multi-vehicle scenarios by MARL, which has shown exciting and outstanding performance in lane changing\cite{wang2021highway,zhang2022multi}, merging\cite{chen2021deep}, intersection\cite{daisocially} scenarios. Nevertheless, these algorithms still fail to guarantee enough security and reliability, which greatly limits further application.
To address these challenges, recent studies have proposed various approaches. For example, some works have incorporated interaction priors in the MARL framework \cite{chen2021deep}. Nevertheless, further research is needed to improve the safety and reliability of MARL algorithms.

Attention mechanism is a cognitive function that is crucial for humans. Recently, this mechanism has been introduced to many fields, including image caption generation, text classification, autonomous driving, and recommendation systems \cite{niu2021review}. It is a newly-emerged technique in neural network models and has shown great power in sequence modeling \cite{vaswani2017attention}. The attention mechanism enables neural networks to identify correlations and inter-dependencies among variable inputs. It has been applied in the tasks of autonomous driving's decision-making, such as capturing vehicle-to-ego dependencies \cite{leurent2019social,zhang2022spatial}, optimizing interactive behavior strategies\cite{daisocially}, and enhancing the safety of the decision-making algorithm\cite{cao2022self,wang2021highway}. Meanwhile, the attention mechanism faces several challenges, including a substantial data requirement \cite{hafiz2021attention}, lengthy training times \cite{niu2021review}, and other issues. Furthermore, the interpretability of the attention mechanism remains to be further explored and enhanced \cite{chaudhari2021attentive}.

\subsection{Summary of Related Works}
The decision-making process for CAVs at intersections is a highly intricate and formidable undertaking. Scholars have proposed various approaches, including game-theoretical models, learning-based techniques, and optimization-based methods, to model decision-making in autonomous driving scenarios, particularly at unsignalized intersections. Notably, Multi-Agent Reinforcement Learning (MARL) has emerged as a promising approach for addressing multi-vehicle interaction and decision-making within complex and densely populated driving environments. However, current MARL algorithms grapple with issues related to non-stationarity, safety, and scalability, thereby curtailing their broader practical application \cite{zhang2021multi}.

Recognizing the potential of the attention mechanism to enhance the safety and reliability of autonomous driving decision-making algorithms, researchers have begun integrating this mechanism into their frameworks. However, it is crucial to acknowledge that existing research on the application of the attention mechanism still presents certain limitations \cite{niu2021review}.

In light of the complex cooperative decision-making challenges faced by CAVs within intricate human-machine mixed traffic scenarios, this study extends the MARL framework by incorporating interaction priors and introducing a safety inspector to address intersection conflicts. Additionally, we explore and leverage the attention mechanism to effectively filter interacting objects in these complex environments.

\section{MARL for Intersection Decision-Making}
\label{section:3}
In this section, the decision-making problem of CAVs is formulated as a MARL problem firstly. To establish a baseline, the MADDPG algorithm for addressing this problem is proposed.

\subsection{Scenario Description and Vehicle Movements}

\subsubsection{Scenario Description}
We consider a cooperative decision-making problem for CAVs in an intersection scenario. A single-lane cross-shaped unsignalized intersection is defined, which may be traversed by a variable number of human-driven vehicles (HVs) arriving from disparate directions and locations, each exhibiting diverse driving styles. In this scenario, four CAVs are introduced, each entering the intersection via dedicated entrance roads corresponding to the four cardinal directions. The objective is for all CAVs to traverse the intersection safely and efficiently, ultimately reaching their respective destinations. Upon the successful completion of this objective, the episode concludes. CAVs can share the status information with each other for better cooperative decision-making.

\subsubsection{Vehicle Movements}
\label{system_dynamic}
The actions of CAVs in our problem are decided by the MARL algorithm and will be translated to low-level steering and acceleration signals through a closed-loop PID controller. The vehicle's position is controlled by
\begin{equation}
    \label{position control}
    \begin{aligned}
        v_{\text{lat},r} &= -K_{p,\text{lat}} \Delta_{\text{lat}}, \\
        \Delta \psi_{r} &= \arcsin \left(\frac{v_{\text{lat},r}}{v}\right)
    \end{aligned}
\end{equation}
The vehicle's heading is controlled by
\begin{equation}
    \label{heading control}
    \begin{aligned}
        \psi_r &= \psi_L + \Delta \psi_{r}, \\
    \dot{\psi}_r &= K_{p,\psi} (\psi_r - \psi), \\
    \delta &= \arcsin \left(\frac{1}{2} \frac{l}{v} \dot{\psi}_r\right)
    \end{aligned}
\end{equation}
where
$\Delta_{\text{lat}}$ is the lateral position of the vehicle with respect to the lane center-line, $v_{\text{lat},r}$ is the lateral velocity command, $\Delta \psi_{r}$ is a heading variation to apply the lateral velocity command, $\psi_L$ is the lane heading, $\psi_r$ is the target heading to follow the lane heading and position, $\dot{\psi}_r$ is the yaw rate command,
$\delta$ is the front-wheel angle control, $K_{p,\text{lat}}$ and $K_{p,\psi}$ are the position and heading control gains.

Vehicle motion is determined by a Kinematic Bicycle Model\cite{polack2017kinematic}
\begin{equation}
\label{bicycle model}
\begin{aligned}
    \dot{x}&=v\cos(\psi+\beta) \\
    \dot{y}&=v\sin(\psi+\beta) \\
    \dot{v}&=a \\
    \dot{\psi}&=\frac{v}{l}\sin\beta \\
    \beta&=\tan^{-1}(1/2\tan\delta) \\
\end{aligned}
\end{equation}
where $(x, y)$ is the vehicle position, $v$ is forward speed, $\psi$ is heading, $a$ is the acceleration command, $\beta$ is the slip angle at the center of gravity.

In this paper, HVs' longitudinal behavior is controlled by the Intelligent Driver Model(IDM) from\cite{kesting2010enhanced}
\begin{equation}
    \label{IDM}
    \begin{aligned}
        \dot{v} &= a\left[1-\left(\frac{v}{v_0}\right)^\delta - \left(\frac{d^*}{d}\right)^2\right] \\
    \end{aligned}
\end{equation}
\begin{equation}
\label{IDM_2}
    d^* = d_0 + Tv + \frac{v\Delta v}{2\sqrt{ab}}
\end{equation}
where $v$ is the vehicle velocity, $d$ is the distance to its front vehicle, $v_0$ is the desired velocity, $T$ is the desired time headway, $d_0$ is the jam distance, $a,\,b$ are the maximum acceleration and deceleration respectively, $\Delta v$ is the velocity exponent.

HVs' lateral behavior is modeled by the Minimizing Overall Braking Induced by Lane change(MOBIL) model\cite{kesting2007general}.

\subsection{Problem Formulation}

The cooperative decision-making problem for multi-CAVs can be formulated as a partially observable Markov decision process (POMDP)\cite{spaan2012partially}. We use the tuple $\mathcal{M}_{\mathcal{G}} = (\mathcal{V}, S,[\mathcal{O}_i],[\mathcal{A}_i],\mathcal{P},[r_i])$ to define the POMDP, in which $\mathcal{V}$ is a finite set of all controlled agents (CAVs) and $S$ denotes the state space describing all agents. $\mathcal{O}_i$ denotes the set of observation spaces for agent $i\in \mathcal{V}$, $\mathcal{A}_i$ is the set of action space, and $r_i$ represents the reward of CAV $i$. $\mathcal{P}$ denotes the transition distribution. Each agent $i $ at a given time receives an individual observation $o_i: \mathcal{S} \to \mathcal{O}_i$ and takes an action $a_i \in \mathcal{A}_i$ based on a policy $\pi_i : \mathcal{O}_i \times \mathcal{A}_i \to [0,1]$. Then the agent $i$ transits to a new state $s_i^\prime $ with probability $\mathcal{P}(s^\prime|s, a): \mathcal{S} \times \mathcal{A}_1 \times ... \times \mathcal{A}_N \to \mathcal{S}$ and obtains a reward $r_i: \mathcal{S} \times \mathcal{A}_i \to \mathbb{R}$.
Each agent aims to maximize its expected return
\begin{equation}
    R_i = \Sigma_{t=0}^{T}\gamma^tr_i(s_t, a_t^i)
\end{equation}
where $\gamma$ represents the discount factor and $T$ represents the time horizon.

\subsubsection{Observation Space}
Due to the limitation of the sensor hardware, the CAV can only detect the status information of surrounding vehicles within a limited distance $\mathcal{L}$. We denote the set of all observable vehicles within the perception range of agent $i$ as $\mathcal{N}_i$. And the observation matrix of agent $i$, denoted as $\mathcal{O}_i$, is a matrix with dimensions of $| \mathcal{N}_i | \times | \mathcal{F} |$, where $|\mathcal{N}_i|$ is the number of all observable vehicles for agent $i$,  and $| \mathcal{F} |$ is to represent the number of features utilized to portray a vehicle's state. The feature vector of the vehicle $k$ is expressed by
\begin{equation}
    \mathcal{F}_k = [x_k, y_k, v^x_k, v^y_k, \cos{\phi_k}, \sin{\phi_k}]
\end{equation}
where $x_k, y_k, v^x_k, v^y_k$ are the longitudinal position, lateral position, longitudinal speed, and lateral speed, $\cos{\phi_k}, \sin{\phi_k}$ are the sine and cosine of the vehicle heading angle $\phi_k$, respectively. 
The whole observation space of the system is the combined observation of all CAVs, i.e., $\mathcal{O} =\mathcal{O}_{1}\times \mathcal{O}_{2}\times \cdots \times \mathcal{O}_{| \mathcal{V} |}$.

\subsubsection{Action Space}
In our research, we pay more attention to the decision-making actions of CAVs rather than the vehicle control level. Specifically, when traversing an intersection, we establish a pre-determined driving route. Within this context, the CAV must make decisions regarding acceleration or deceleration in order to execute a left turn at the intersection and ultimately arrive at its intended destination.
After selecting a high-level decision, lower-level controllers generate the corresponding steering and throttle control signals to control the CAVs' movement.  The overall action space is the combined actions of all CAVs, i.e., $\mathcal{A} =\mathcal{A}_{1}\times \mathcal{A}_{2}\times \cdots \times \mathcal{A}_{| \mathcal{V} |}$.

\subsubsection{Reward Function}
The reward function has a great effect on the final performance of the algorithm. In order to make agents pass the intersection safely and effectively, the reward of $i$th agent at the time step $t$ is defined as
\begin{equation}
    r_{i,t} = \underbrace{w_c r_c}_{\text{Collision reward}} + \underbrace{w_e r_e}_{\text{Efficiency reward}} + \underbrace{w_a r_a}_{\text{Arrival reward}}
\end{equation}
where $w_c$,$w_e$, and $w_a$ are the weight coefficients of collision reward $r_c$, efficiency reward $r_e$, and arrival reward $r_a$, respectively. These evaluation terms are defined as follows:
\begin{itemize}
    \item Collision reward $r_c$: Safety is the most important criterion for a vehicle. In order to make the agent learn to drive safely,  we give it a greater penalty when a collision happens.
    \begin{equation}
        r_c =  \left\{
                     \begin{array}{lr}
                     -10 \quad if \ agent_i \ collide  \\
                     0 \quad otherwise
                     \end{array}
        \right.
    \end{equation}
    \item Efficiency reward $r_e$: The speed range $[v_{min}, v_{max}]$ for the agents passing the intersection is set based on real-world traffic rules. The agent is not recommended to drive at a speed outside this range, and we encourage the agent to pass as efficiently as possible within the speed range. And a constant $C_e$ is used to adjust the maximum speed bonus.
    \begin{equation}
        r_e = C_e \times \min{\Big\{\frac{v_i-v_{min}}{v_{max}-v_{min}}\Big\}}
    \end{equation}
    \item Arrival reward $r_a$: When any agent completes its ultimate goal to reach the end, they will obtain an arrival reward.
    \begin{equation}
        r_a =  \left\{
                     \begin{array}{lr}
                     5 \quad if \ agent_i \ reaches \ destination  \\
                     0 \quad otherwise
                     \end{array}
        \right.
    \end{equation}
\end{itemize}

where $T$ is the total time steps for one episode.
\begin{figure}[!htbp]
    \centering
    \includegraphics[width=0.5\textwidth]{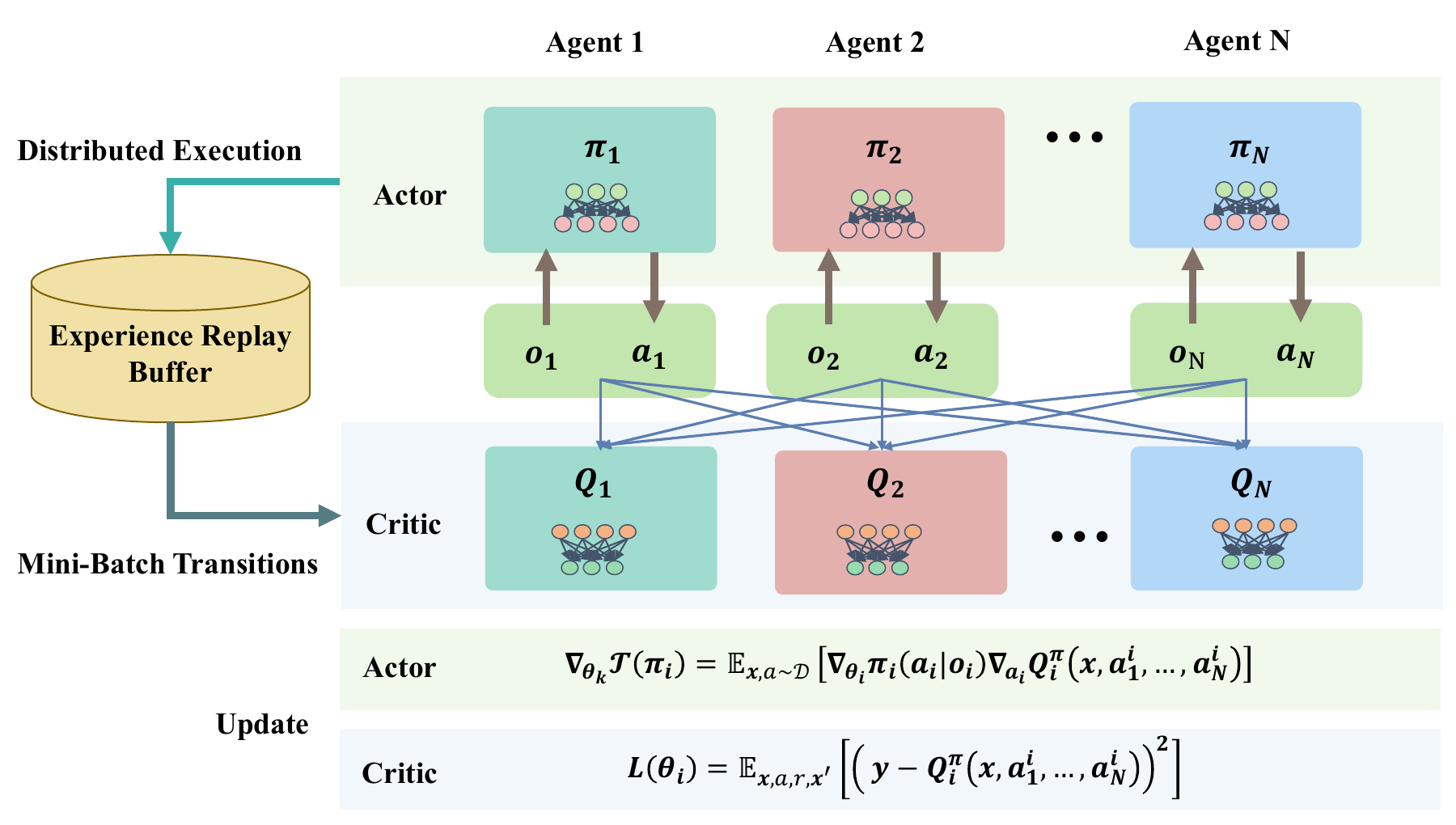}
    \caption{The framework of the Multi-Agent Deep Deterministic Policy Gradient (MADDPG).}
    \label{fig:MADDPG}
\end{figure}

\begin{figure*}[!htbp]
    \centering
    \includegraphics[width=0.8\textwidth]{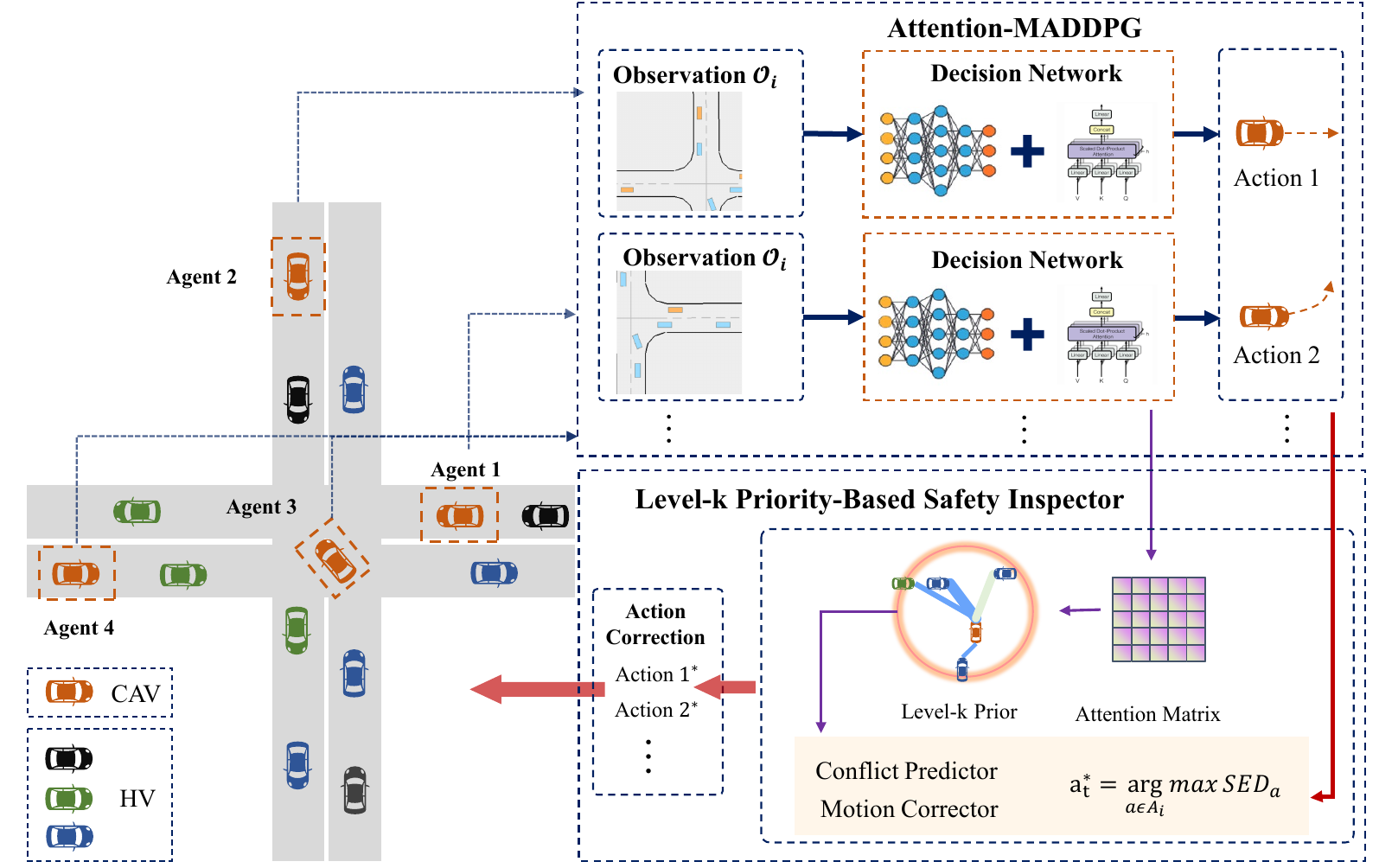}
    \caption{The overview of our MA-GA-DDPG Model.}
    \label{fig:overview_framework}
\end{figure*}

\subsection{MADDPG For Decision-Making of CAVs}
In our work, a model-free MARL algorithm, MADDPG, is utilized as the baseline algorithm.

In Deep Reinforcement Learning (DRL), a deep neural network (DNN) serves as a non-linear approximator to obtain optimal policies $\pi^{*}$ for CAVs (agents). The agents interact with the environment and receive feedback in the form of rewards, which are used to update the agent's policy. Let $\pi = \big\{\pi_{1}, \cdots, \pi_{N}\big\}$ denote the set of all CAVs' policies and $\theta = \big\{\theta_{1}, \cdots, \theta_{N}\big\}$ denote the parameter set of corresponding policy, where $N =| \mathcal{V} | $ is the number of CAVs. CAVs update their policies based on the estimation of the Q-function for each possible action using the off-policy actor-critic algorithm MADDPG~\cite{lowe2017multi}. The objective function for MADDPG is the expected reward $\mathcal{J}(\theta)$, i.e., $\mathcal{J}(\theta_{i}) = \mathbb{E}[\Omega_{i}(t)]$. The optimal policy of each CAV is represented as $\pi_{\theta_{i}}^{*} = \arg \max_{\pi_{\theta_{i}}}\mathcal{J}(\theta_{i})$. Then the algorithm calculates the gradient of the objective function with respect to $\theta_{i}$ as
\begin{equation}
    \label{eq:gradient_of_Objective}
    \nabla_{\theta_{i}}\mathcal{J}(\pi_{i}) = \mathbb{E}_{\textbf{x}, a \thicksim \mathcal{D}}[\nabla_{\theta_{i}} \pi_{i}(a_{i}|o_{i})\nabla_{a_{i}}Q_{i}^{\pi}(\textbf{x}, a_{1}, \cdots, a_{N})],
\end{equation}
where $\textbf{x} = \mathcal{O} = (o_{1}, \cdots, o_{N})$, $Q_{i}^{\pi}(\textbf{x}, a_{1}, \cdots, a_{N})$ is a centralized action-value funciton, and $\mathcal{D}$ is the replay buffer. $\mathcal{D}$ contains transition tuples $(\textbf{x}, a, r,  \textbf{x}^{\prime})$, where $a = (a_{1}, \cdots, a_{N})$ and $r = (r_{1}, \cdots, r_{N})$. To minimize the loss function \eqref{eq:para_update}, the centralized action-value function $Q_{i}^{\pi}$ is updated for 
\begin{equation}
\label{eq:para_update}
    \mathcal{L}(\theta_{i}) = \mathbb{E}_{\textbf{x}, a, r, \textbf{x}^{\prime}}[(y - Q_{i}^{\pi}(\textbf{x}, a_{1}, \cdots, a_{N}))^2],
\end{equation}
where $y = r_{i} + \gamma Q_{i}^{\pi^{\prime}}(\textbf{x}^{\prime}, a_{1}^{\prime}, \cdots, a_{N}^{\prime})|_{a_{j}^{\prime}=\pi_{j}^{\prime}(o_{j})}$. $\pi^{\prime} = \big\{\pi_{\theta_{1}^{\prime}}, \cdots, \pi_{\theta_{N}^{'}}\big\}$ stands for the target policies with delayed parameters $\theta_{i}^{'}$. The schematic diagram is shown in Fig. \ref{fig:MADDPG}

\section{Game-prior Attention Model}
\label{section:4}
In this section, the whole framework of our MA-GA-DDPG model is first outlined. Then we introduce an attention mechanism-based policy network and an interaction filter approach. Furthermore, a safety inspector based on the attention weights and the level-k game is applied to enhance the safety performance of the algorithm.
\subsection{Model Overview}

The schematic representation of our model is depicted in Figure \ref{fig:overview_framework}. Initially, we devise an innovative policy network employing the attention mechanism for each agent within the MADDPG algorithm. This network adopts an encoder-decoder framework, incorporating a multi-head attention layer to comprehensively capture interaction relationships among agents. Subsequently, attention weights for each Connected and Autonomous Vehicle (CAV) in relation to all surrounding traffic participants are computed based on the attention-based policy network. We then introduce a filtering rule employing these attention weights to effectively screen strongly interacting participants.

Simultaneously, the attention degree acquired by the policy network for each vehicle in the scenario is construed as the priority weight for the vehicle's passage through the intersection. Leveraging the concept of a level-k game (hierarchical game), these priority weights are transformed into a level prior. Subsequently, a safety inspector is defined to oversee the safety aspect. Utilizing the input game prior, the safety inspector proficiently anticipates, evaluates, and rectifies high-risk actions in a timely manner, thereby mitigating or resolving conflicts for each CAV. The subsequent presentation of the safety inspector demonstrates its efficacy in enhancing the safety of the RL algorithm.

\subsection{Attention-Based Policy Network}

The attention mechanism has been well-documented to enable neural networks to discover interdependencies among a variable number of inputs and has been applied in the social interaction research of autonomous vehicles\cite{leurent2019social,daisocially}.

\begin{figure}[!htbp]
    \centering
    \includegraphics[width=0.5\textwidth]{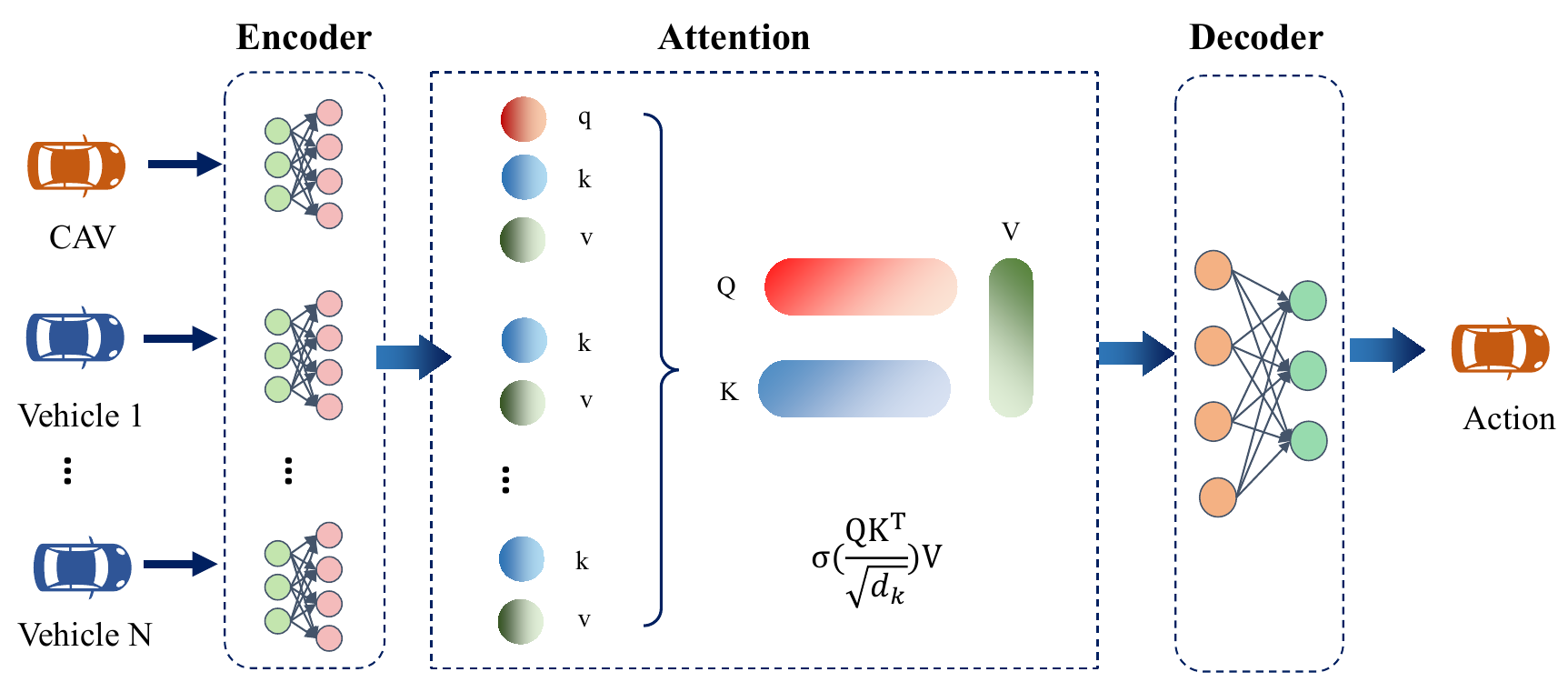}
    \caption{The attention-based policy network for every single agent.}
    \label{fig:attention-based policy network}
\end{figure}

Inspired by \cite{leurent2019social}, we design an attention-based policy network for each agent in a decentralized training process, as shown in Fig. \ref{fig:attention-based policy network}. The network contains three modules: encoder block, attention block, and decoder block. 
In the encoder block, the features $\mathcal{F}_i$ of the agent $i$ and its observation matrix $\mathcal{O}_i$ are encoded as high-dimension vectors by a Multilayer Perceptron (MLP), whose weights are shared between all vehicles. 
\begin{equation}
\label{eq:state_encoding}
    \mathcal{X}_i = MLP(\mathcal{F}_i,\mathcal{O}_i)
\end{equation}

And then the feature matrix is fed to the attention block, composed of $N_{head}$ attention heads stacked together. Unlike the attention layer in the Transformer model\cite{vaswani2017attention}, this block produces only the query results (attention weights) of agent $i$, which indicate how much attention it should pay to the surrounding vehicles.

In the attention block, the ego vehicle emits a single query $Q_i=[q_0]\in \mathbbm{R} ^{1\times d_k}$ to select a subset of vehicles based on the environment, where $d_k$ is the output dimension of the encoder layer. This query is then projected linearly and compared to a set of keys $K_i=[k^0_i,k^1_i, \cdots,k^N_i]\in R^{(N+1)\times d_k} $ containing descriptive features for each vehicle, using dot product $q_0 k_i^T$ to calculate the similarity. The $Q_i$, $K_i$ and $V_i$ are calculated as follows.
\begin{equation}
\label{eq:attention_matrix_mapping}
    \begin{aligned}
        & Q_i = W^Q \mathcal{X}_i \\
        & K_i = W^K \mathcal{X}_i \\
        & V_i = W^V \mathcal{X}_i
    \end{aligned}
\end{equation}
where the dimensions of $W^Q$ and $W^K$ are $(d_k\times d_N)$, and $W^V$'s is $(d_v \times d_h)$.

The attention matrix is obtained by scaling the dot product with the inverse-square-root-dimension $\frac{1}{\sqrt{d_k}}$ and normalizing it with a softmax function $\sigma$. The attention matrix is then used to gather a set of output values $V_i=[v^0_i,\cdots,v^N_i]$, where each $v^j_i$ is a feature computed using a shared linear projection $L_v \in \mathbbm{R}^{d_x \times d_k}$. The attention computation for each head can be written as
\begin{equation}
\label{eq:cal_attention_1}
    At_i^m = \sigma \Big (  \frac{Q_i K^T_i}{\sqrt{d_k}} \Big)V_i
\end{equation}

Then the output from all $M$ heads will be combined with a linear layer:
\begin{equation}
\label{eq:cal_attention_2}
    At_i = \sum_{m=1}^{M} At_i^m
\end{equation}

The final attention vector is denoted by $At_i=\big[At_{i,1}, At_{i,2},\cdots, At_{i,|\mathcal{N}_i|} \big]$, where $At_{i,j} (j \in \mathcal{N}_i)$ denotes the weight of the agent $i$'s attention to the surrounding vehicle $j$, satisfying the cumulative summation relationship: $\sum^{|\mathcal{N}_i|}_{j=1} At_{i,j} = 1$.  The vector $At_i$ will be fed to the decoder block along with the encoding result of agent $i$. Finally, the value of the agent $i$’s action at the next time step will be assessed.

Overall, the attention-based policy network has the following significant advantages: (1) It can handle the variable amount of observation information inputs; (2) It has permutation invariant outputs that are independent of the sequence of surrounding agents; (3) It has good interactive interpretability based on the attention matrix.
In the next subsection, we will screen the interactive objects based on the interpretable feature of the attention mechanism and obtain the intersection traffic prior based on the hierarchical game.

\subsection{Hierarchical Game Prior From Agent's Attention}
The interpretability of the attention network allows us to further utilize the learned attention weight information. In the real world, in order to pass through intersections, both CAVs and HVs will have competitive or cooperative game relationships with each other. If this game relationship can be learned and sent to the algorithm as prior information, it will greatly help the algorithm explore and learn more efficiently.

To capture the strategic decision-making process of HVs, a game-theoretical concept named level-k reasoning is used~\cite{li2020game}. This approach assumes that humans have different levels of reasoning, with level-0 being the lowest. A level-0 agent is a non-strategic agent that makes predetermined moves without considering the possible actions of other agents. On the other hand, a level-1 agent is a strategic agent that assumes all other agents have level-0 reasoning and decides the best response to such actions. Similarly, a level-2 agent assumes all other agents are level-1 and makes decisions based on this assumption. This hierarchy continues for higher levels. However, due to bounded rationality for the agents\cite{wen2021modelling}, this assumption may not always hold true. The experiments reveal that humans generally have at most level-3 reasoning\cite{yuan2021deep}, but this may vary depending on the game being played.

\begin{figure*}[!htbp]
    \centering
    \includegraphics[width=0.9\textwidth]{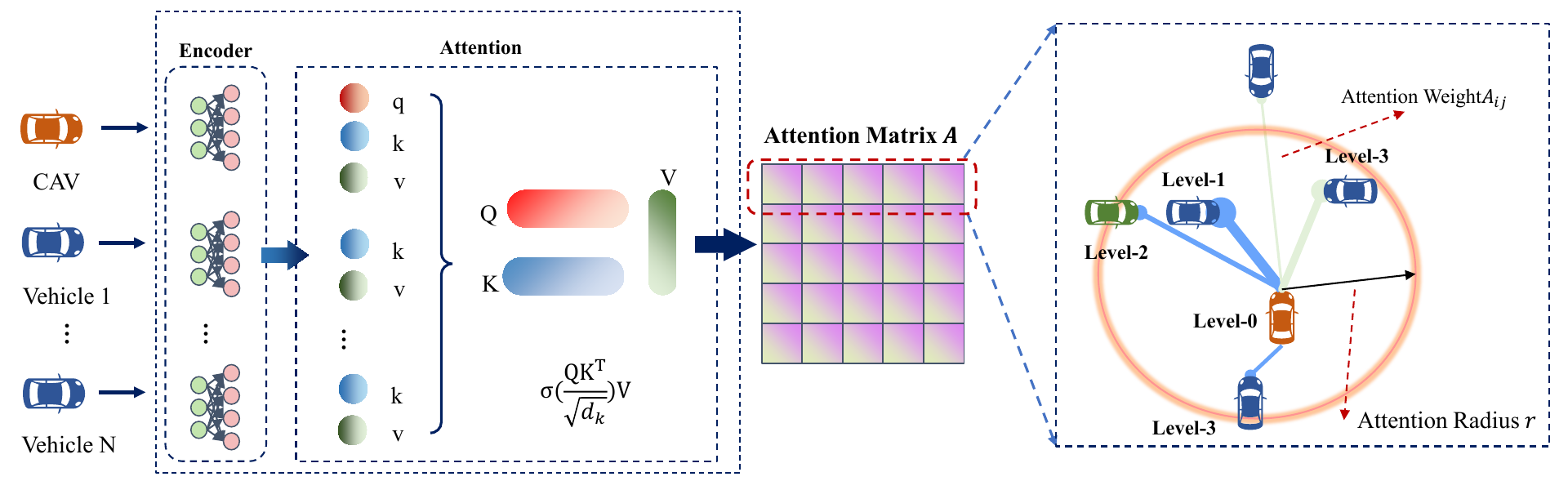}
    \caption{Attention-based interactive object selection for each CAV.}
    \label{fig:attention_interactive_object_selection}
\end{figure*}

The level of the agent is obtained based on the attention mechanism and is considered as important prior knowledge. As shown in Fig. \ref{fig:attention_interactive_object_selection}, our method mainly includes two steps: Interactive Object Selection and Level Prior Determination. The detailed process is summarized in Algorithm \ref{algo:interaction_objects_selection}.

\begin{itemize}
    \item $Step \ 1$ : Interactive Object Selection.
    Similar to human attention, we set an attention radius $dis_0$ and an attention weight threshold $\delta_0$ for each CAV to select potential interaction objects. At each moment, when a surrounding vehicle $j$ satisfies the condition that the distance $dis_{ij}$ between $j$ and CAV $i$ is less than $dis_0$ and the attention weight $At_{i,j}$ is greater than $\delta_0$, and the interaction limit $\mathcal{Q}$ has not been reached, it is included in the set of Potential interaction Objects $PO_{inter}$.
    \item $Step \ 2$ : Level Prior Determination.
    The interaction importance is determined based on the attention weight. In the interaction environment, a vehicle that receives more attention is believed to be more important to the CAV, and vice versa. In an environment with only one CAV, the interaction importance of the environment vehicle with the highest attention weight (i.e., the most attention received) is ranked highest, followed by the others in descending order of attention weight. Finally, we obtain a list of interaction importance $Rank_i$ between the CAV and all surrounding vehicles. In a multi-CAVs environment, since each CAV can calculate the attention weight for all vehicles, the global attention weight of vehicle $j$ is the sum of the attention weight given by all CAVs:
    \begin{equation}
        BAt^j = \sum_{i=1}^{\mathcal{V}} At_{i,j}
    \end{equation}
    The environment vehicle with the highest attention weight and priority is ranked highest, followed by the others in descending order of attention weight. If the interaction object limit $\mathcal{Q}$ of the CAV is reached, the top $Q$ objects in the sorted set $PO_{inter}$ are selected.
\end{itemize}

\begin{algorithm}
\SetAlFnt{\small}
    \SetKwInOut{Parameter}{Inputs}
    \SetKwInOut{Output}{Outputs}
\caption{Obtain Hierarchical Level Priors Based On Attention Mechanism}
\label{algo:interaction_objects_selection}
\LinesNumbered 
\SetAlgoLined
\Parameter{$At, \ dis_0, \ \delta_0, \ \mathcal{Q}$}
\Output{ $PO_{inter}, \ Rank$}
\vspace{0.2em}
\hrule
\vspace{0.2em}
\textbf{Step 1:}\\
\For{$i = 1$ to $| \mathcal{V} | (i \in \mathcal{V})$}{
    Calculate Attention vector $At_i$ by Eq.\ref{eq:cal_attention_1} and Eq.\ref{eq:cal_attention_2}; \\
    \For{$j = 1$ to $|\mathcal{V} + \mathcal{N}_i|\ (j \in |\mathcal{V}|\cup \mathcal{N}_i)$}
    {
    Calculate the Euclidean Distance between agent $i$ and $j$ : $Dis(i,j)$;\\
    \If{$Dis(i,j) > dis_0$ and $At_{i,j} > \delta_0$}
    {
    $PO^i_{inter} = PO^i_{inter} \cup j$ \\
    \If{$| PO^i_{inter}| > \mathcal{Q}$}
    {
    Remove $PO^i_{inter} [-1]$ from $PO^i_{inter}$;\\
    }
    }
    Sort $PO^i_{inter}$ in descending order based on $At_{i,j}$;\\
    }
    }
\textbf{Step 2:} \\
\For{$j = 1$ to $|\mathcal{V} + \mathcal{N}_i|(j \in |\mathcal{V}|\cup \mathcal{N}_i)$}
{
    $BAt_j = \sum^{\mathcal{V}}_{i=0} \sigma At_{i,j}$; \\
    $\sigma =  \left\{
                 \begin{array}{lr}
                 1 \  if \ j \ \in PO^i_{inter}  \\
                 0 \  else
                 \end{array}
    \right. $
}
Sort $BAt$ in descending order;\\
\For{$j = 1$ to $|\mathcal{V} + \mathcal{N}_i|\ (j \in |\mathcal{V}|\cup \mathcal{N}_i)$}
{
Update:
$Rank_{i,j} = Index(BAt_j)$;
}
\end{algorithm}

\subsection{Safety Inspector Based On Hierarchical Game Prior}
During vehicular passage through an intersection, the prevalent assumption is that a vehicle garnering more attention is considered to hold greater significance or pose an increased degree of risk. Consequently, we contend that vehicles characterized by higher attention weights should be afforded a higher priority or right of way at intersections. This established right-of-way hierarchy can serve as foundational information to expedite the learning of action strategies by the RL algorithm that align with real-world logical constructs. With this objective in mind, we have architected a safety inspector module based on the level-k priority principle. Analogous to the $Critic$ network within the RL algorithm, this safety inspector module operates independently of the policy network.

\begin{figure}[!htbp]
    \centering
    \includegraphics[width=0.4\textwidth]{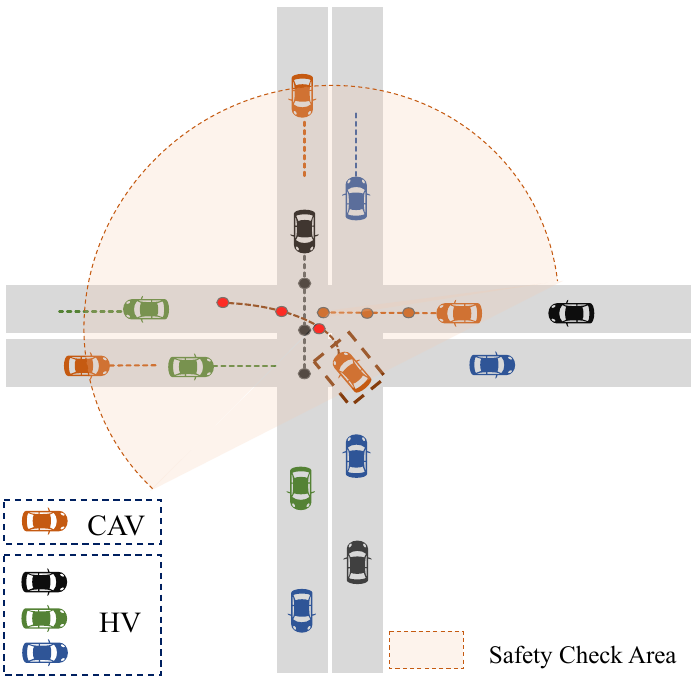}
    \caption{Trajectory prediction of surrounding agents and conflict checking for CAV $i$ at the intersection}
    \label{fig:safey_checker_at_intersection}
\end{figure}

The safety inspector module functions as an active conflict regulator, embodying two primary components: the Proactive Predictor of Agent Risk and the Assisted Corrector of Agent Motion. These elements are instrumental in assisting and rectifying the agent's exploratory behavior during the initial stages, thereby ensuring the algorithm's performance maintains continuity and stability. The safety supervisor, founded on the level-k priority-based framework, is visually represented in Figure \ref{fig:safey_checker_at_intersection}.

\begin{algorithm}
\SetAlFnt{\small}
\SetKwInOut{Parameter}{Inputs}
\SetKwInOut{Output}{Output}
\caption{Level-k Priority-based Safety Inspector}
\label{algo:safety_inspector}
\LinesNumbered 
\SetAlgoLined
\Parameter{$t_0$, $T$, $A_{t_0}$, $O$, $S$}
\Output{$A^*_{t_0}$}
\vspace{0.2em}
\hrule
\vspace{0.2em}
\For{$i = 1$ to $|\mathcal{V}| \ (i \in \mathcal{V})$}
{
    Obtain the output action $a_{t_{0}}$ of policy network and level Prior $k_i$ by Alg. \ref{algo:interaction_objects_selection}; \\
    \For{$t = 1$ to $T$}
    {
    Sample future trajectories based on $a_{t_0}$ by $MDP$:
    $(x^i_{k+1}, y^i_{k+1}) = MDP(x^i_k, y^i_k)$
    }
    $\big( tra_{t_0}^i \big)_{a_{t_0}}^{k_i} = \left\{ \left(x_{t_0 + 1}^i, y_{t_0 + 1}^i \right)^{k_i}, \ldots, \left(x_{t_0 + T}^i, y_{t_0 + T}^i \right)^{k_i}\right\}$;\\
    \For{$j = 1$ to $|\mathcal{V}| \ (j \in \mathcal{V} \ and \ j \neq i)$}
    {
    Get $\big( tra_{t_0}^j \big)^{k_j}$ by V2V communication;
    }
    \For{$m = 1$ to $|\mathcal{N}_i| \ (m \in \mathcal{N}_i)$}
    {
    Predict future trajectories based on $\mathcal{O}_i$ by $IDM$:
    $\big( tra_{t_0}^m \big)^{k_m} = IDM(x^m_k, y^m_k) $
    }
    Calculate the number of conflict points $CI(a, Tra) \ (a \in A_{t_0})$;\\
    \For{$a^{\prime} \in \mathcal{A}_i$}
    {
    Calculate $CI(a^{\prime}, Tra)$;\\
    Get $SED(a, a^{\prime}, Tra)$ by Eq. \ref{eq:SED};
    }
    Solve Eq.\ref{eq:action_correction} and get $a_{t_0}^{i *}$;\\
    Update $A^*_{t_0} = A^*_{t_0} \cup a_{t_0}^{i *}$;
}
\end{algorithm}

The specific calculation process of the safety inspector module is as follows:
\begin{itemize}
    \item $Step \ 1$: Agent Trajectory Prediction and Conflict Checking.
    Through the communication between CAVs, the safety inspector obtains the state information O of all perceived vehicles in the intersection at any time step $t_0$, and the current action decision set $\mathcal{A}_{t_0}$ of all CAVs. Then based on the level-k prior information, it selects agent $i$ from the CAV set $\mathcal{V}$ in descending order to carry out active prediction of agent risk. For agent $i$, let $k_i$ denote its prior level, and $T$ denote the  forward prediction step size. Obtain the future trajectory sequences of agent $i$ :
    \begin{equation}
    \begin{aligned}
        & \big( tra_{t_0}^i \big)_{a_{t_0}}^{k_i} = \big \{ \big(x_{t_0 + 1}^i,y_{t_0 + 1}^i \big)^{k_i},
        \big(x_{t_0 + 2}^i,y_{t_0 + 2}^i \big)^{k_i}, \cdots,\\
        & \big(x_{t_0 + T}^i,y_{t_0 + T}^i \big)^{k_i}\big \}(a_{t_0} \in \mathcal{A}_{t_0})
    \end{aligned}
    \end{equation}
    by sampling T steps through the MDP process, and obtain the future $T$ step trajectories of other CAV $j$  through communication:
    \begin{equation}
    \begin{aligned}
        & \big( tra_{t_0}^j \big)^{k_j} = \big \{ \big(x_{t_0 + 1}^j,y_{t_0 + 1}^j \big)^{k_j},\big(x_{t_0 + 2}^j,y_{t_0 + 2}^j \big)^{k_j}, \cdots,\\
        & \big(x_{t_0 + T}^j,y_{t_0 + T}^j \big)^{k_j}\big \} (j\in  \mathcal{V})
    \end{aligned}
    \end{equation} 
    And for all HV $m$ within the observation range of $i$, use the IDM model to predict their future trajectories :
    \begin{equation}
    \begin{aligned}
        & \big( tra_{t_0}^m \big)^{k_m} = \big \{ \big(x_{t_0 + 1}^m,y_{t_0 + 1}^m \big)^{k_m},
        \big(x_{t_0 + 2}^m,y_{t_0 + 2}^m \big)^{k_m}, \cdots,\\
        & \big(x_{t_0 + T}^m,y_{t_0 + T}^m \big)^{k_m}\big \} (m \in  \mathcal{N}_i)
    \end{aligned}
    \end{equation} 
    Then judge whether there is a conflict between the future trajectories of agent $i$ and all other vehicles. If there is a conflict, jump to $Step \ 2$, otherwise, the output action of the agent is safe, and the safety check of agent $i$ at time step $t$ ends.
    
    \item $Step \ 2$: Action Correction For Agent's Decision-making.
    If the future trajectory of agent $i$ conflicts with other vehicles, the optimal alternative action needs to be found to minimize the conflict in the intersection. We use the number of conflict points of all vehicles in the scenario as the conflicting index $CI$ to measure the danger degree of  the scenario at each timestamp. Based on the conflicting index $CI$, we define the Safety Enhancement Degree $SED$ function to measure the degree of conflict mitigation brought about by agent action correction:
\begin{equation}
\label{eq:SED}
    \begin{aligned}
            SED(a,a^{\prime},Tra)= CI(a,Tra) - CI(a^{\prime},Tra)
    \end{aligned}
\end{equation}
where $Tra$ is the future trajectories set of all vehicles, $a$ and $a^{\prime}$ are the origin action and corrected action of agent $i$, respectively, and $Tra^{a^{\prime}}$ is the future trajectories set of all vehicles based on the corrected action $a^{\prime}$.

So our objective function for agent $i$ is derived by
\begin{equation}
\label{eq:action_correction}
    \begin{aligned}
        & a_{t_0}^{i *} =  \underset{a^{\prime}\in \mathcal{A}_i}{\arg} \max SED \Big(a, a^{\prime}, (tra_{t_0}^i \big)^{k_i}, (tra_{t_0}^j)^{k_j},\\
        & \cdots, (tra_{t_0}^m )^{k_m} \Big)
    \end{aligned}
\end{equation}

    \item $Step \ 3$: Output the optimal actions.\\
    Judging whether all agents have completed the above process, if all have been completed, execute all agent-corrected security actions $\mathcal{A}_{t_0}^*$.
\end{itemize}

The operational procedure of the safety inspector is algorithmically described in Algorithm \ref{algo:safety_inspector}.

Much like the MADDPG algorithm, the MA-GA-DDPG algorithm is grounded in the actor-critic model. Within this framework, the actor assumes the role of decision-maker over time, while the critic evaluates the actor's behavior. Each agent within the system is equipped with both an actor and a critic, encompassing behavior and target networks. 
CAVs strive to enhance their own policy to maximize rewards, simultaneously updating their critic's Q-function to evaluate actions effectively. The primary objective is to adapt the target network's parameters $\theta$ for CAVs to learn optimal action strategies.
The comprehensive representation of our refined model is encapsulated in Algorithm \ref{algo:marl_safety}.

\begin{algorithm}
\SetAlFnt{\small}
    \SetKwInOut{Parameter}{Inputs}
    \SetKwInOut{Output}{Output}
\caption{MA-GA-DDPG for CAVs}
\label{algo:marl_safety}
\LinesNumbered
\SetAlgoLined
\Parameter{$T_{Max}, M, dis_0, \delta_0, \mathcal{Q}$}
\Output{$\theta$}
\hrule
\vspace{0.2em}
\For{$episode = 1$ to $M$}
{
{\bf Initialize} $\mathcal{D} \leftarrow \emptyset$, a random process $\mathcal{G}$ for action exploration;\\
Receive initial state $\textbf{x}$;\\

    \For{$t=1$ to $T_{max}$}
    {
        \For{$CAV \ i = 1 \in \mathcal{V} $}
        {
            Get observation $o_{i,t}$;\\ 
            Update action $a_i=\pi_{\theta_i}(o_{i,t})+\mathcal{G}_t$;
        }
            
        \For{$i = 1 \in \mathcal{V}$}
        {
            
            Get attention weights $At_i$ by Alg. \ref{algo:interaction_objects_selection};\\
            Get corrected actions $a^*_{i,t}$ by $At_i$ and Alg.~\ref{algo:safety_inspector};\\
            Execute $a^{*}_{i,t}$ and update $a_{i,t} \leftarrow a^{*}_{i,t}$;\\
            Observe reward $r_{i,t}$ and new observation $\textbf{x}^{\prime}_{i,t}$;\\
            Update $\mathcal{D}_i \leftarrow (\textbf{x}_{i,t}, a^{*}_{i,t}, r_{i,t}, \textbf{x}^{\prime}_{i,t})$;\\
        }
        
        $\mathbf{x}\gets\mathbf{x}^{\prime}$;\\
        
        \For{ CAV $i=1 \in \mathcal{V}$}
        {
            Sample a random minibatch of $S$ samples $(\textbf{x}_{i,t}, a^{*}_{i,t}, r_{i,t}, \textbf{x}^{\prime}_{i,t})$ from $\mathcal{D}_i$; \\
            
            Set $y^j=r_i^j+\gamma Q_i^{\pi ^{\prime}}(\mathbf{x}'^j,a_{1,t}',\ldots,a_{N,t}')$;\\
            Update critic by minimizing the loss  
            $\mathcal{L}(\theta_i)=\frac{1}{S}\sum_j\left(y^j-Q_i^{\pi}(\mathbf{x}^j,a^j_1,\ldots,a^j_N)\right)^2$; \\
            Update actor using the sampled policy gradient: \\
            $
            \nabla_{\theta_i}(\pi_i) \approx\frac{1}{S}\sum_j\nabla_{\theta_i}\pi_i(o_i^j)\nabla_{a_i}Q_i^{\pi}(\mathbf{x}^j,a^j_1,
            \ldots,a^j_N)
            $
        }
        Update target network parameters for each CAV $i$:\\
          $\theta_i^{\prime}\gets\tau\theta_i+(1-\tau)\theta_i^{\prime}$
    }
}
\end{algorithm}

\section{Simulation and Performance Evaluation}
\label{section:5}
This section commences by introducing the simulation environment, providing an overview of the defined environment, followed by a detailed description of the training experiments and associated hyperparameters. Subsequently, the experimental results are presented, and a thorough analysis of various cases is conducted.

\subsection{Simulation Environment}
\label{simulation_env}

Building upon the foundation of an OpenAI Gym environment\cite{highway-env}, we have formulated a RL training simulator conducive to multi-agent centralized training and distributed execution. The simulator currently offers the flexibility to tailor the intersection environment to specific requirements, accommodating both CAVs and HVs. It is designed to extend its applicability to diverse scenarios beyond intersections. Within this simulator, actions prescribed by specific policies undergo translation into low-level steering and acceleration signals via a closed-loop PID controller. Additionally, the longitudinal and lateral decisions of HVs are governed by the IDM and the MOBIL model, respectively, as explicated in Section \ref{system_dynamic}.

Moreover, to emulate the perception and prediction capabilities of human drivers regarding the movement of other traffic participants, all HVs are endowed with constant-speed motion prediction and collision avoidance functions looking $T_h$ seconds into the future. For each simulation episode, we introduce randomized initial states for all agents, thereby preventing the policy network from memorizing specific action sequences and instead fostering the acquisition of generalized policies.
\subsection{Simulation Settings}

\subsubsection{Training Scenarios Design}
We have designed a mixed human-machine environment to rigorously evaluate our algorithm. To enhance the realism of the environment, we have taken into account the heterogeneity of drivers, incorporating three distinct driving styles: Aggressive, Normal, and Timid, as defined by Zhang et al. \cite{zhang2021comprehensive}. The corresponding parameters for each driving style are outlined in Table \ref{tab:Driving_style_parameter}.

\begin{table}[htp]
\caption{The parameters of different driving styles.}
\label{tab:Driving_style_parameter}
\centering
\resizebox{\linewidth}{!}
{
\begin{tabular}{@{}ccccc@{}}
\toprule
Driving Style   & 
\begin{tabular}[c]{@{}c@{}} $Jam Distance$ \\ $(d_0)(m)$\end{tabular} & 
\begin{tabular}[c]{@{}c@{}} $Desired Time $ \\ $Headway(T)(s)$\end{tabular} &
\begin{tabular}[c]{@{}c@{}} $Maximum$ \\ $Acceleration(a_0)(m/s^2)$\end{tabular} & \begin{tabular}[c]{@{}c@{}}$Maximum$ \\$Deceleration(b_0) (m/s^2)$ \end{tabular}\\
\midrule
        Aggressive & 3.38 & 0.86 & 1.35 & 2.07 \\
        Normal & 3.67 & 1.14 & 1.34 & 2.06 \\
        Timid & 3.69 &1.27 &1.36 & 1.99 \\
\bottomrule
\end{tabular}
}
\end{table}

Overall, we have configured three progressively challenging scenarios:
\begin{enumerate}
\item [(a)] Scenario with only CAVs, each entering from a separate lane.
\item [(b)] Scenario with four CAVs and a varying number of homogeneous HVs, wherein all HVs exhibit the same driving style.
\item [(c)] Scenario with four CAVs and a varying number of heterogeneous HVs, wherein each HV possesses a distinct randomly generated driving style.
\end{enumerate}

In addition, we have employed two baseline algorithms, namely MADDPG and Attention-MADDPG (MADDPG Algorithm with attention-based policy network), for comparative evaluation.

\begin{figure*}[!htbp]
    \centering
    \includegraphics[width=1\textwidth]{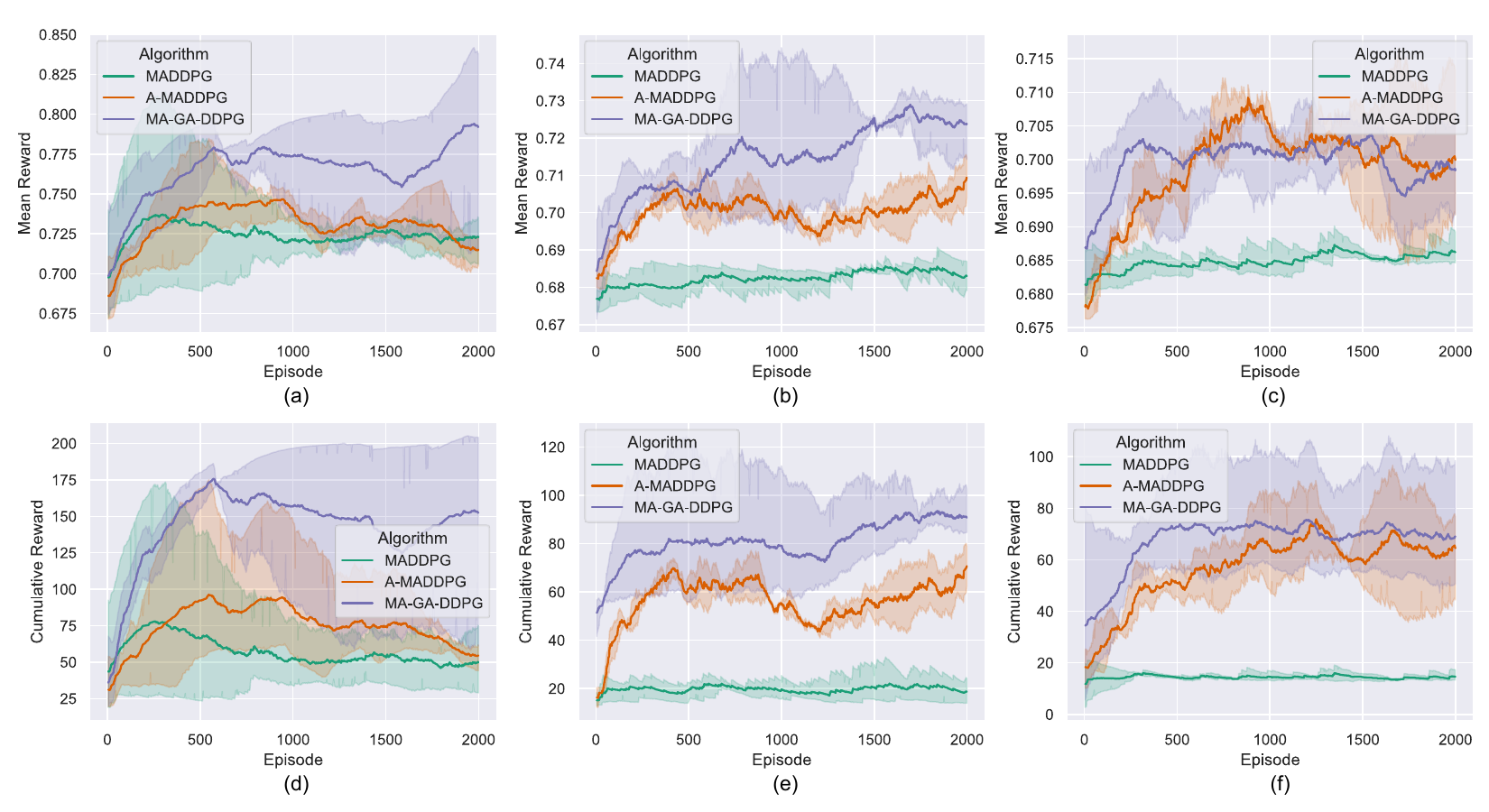}
    \caption{The mean reward and cumulative reward of our model and other baselines in different environments, (a) and (d): just CAVs; (b) and (e): CAVs and homogeneous HVs; (c) and (d): CAVs and heterogeneous HVs.}
    \label{fig:alg_results}
\end{figure*}

\subsubsection{Implementation Details}
The hyperparameter of the model during the training is shown in Table \ref{tab:training_hyperparameter}. The speed range: $[v_{min}, v_{max}]$ is $[3.0, 9.0]$. In the attention-based policy network, the Encoder and Decoder both are MLP, which has two linear layers and the layer size is $64 \times 64$. The Attention Layer contains two heads and the feature size is set as 128. When selecting the interaction with attention weights, we set $dis_0 = 40, \delta_0 = 0.5$, and $\mathcal{Q}=5$. In the safety inspector module, we predict $T = 5$ steps feature trajectories for each vehicle. All the experiments are conducted in a platform with Intel Core i7-12700 CPU, NVIDIA GeForce RTX 3070 Ti GPU, and 32G memory.

\begin{table}[!htbp]
    \centering
    \caption{The hyperparameter of the model for training.}
    \label{tab:training_hyperparameter}
    \begin{tabular}{c c c}
        \toprule
        Symbol & Definition & Value\\
        \midrule
        $N_t$ & Training Episodes & 2000 \\
        $S_u$ & Steps Per Update & 100 \\
        $\Psi$ & Buffer Length & 10000 \\
        $\lambda$ & Learning Rate & 0.01 \\
        $B$ & Batch of Transitions & 128\\
        $\gamma$ & Discount factor & 0.95\\
        $\tau$ & Target update rate	& 0.01\\
        $w_c$ & Weight for $r_c$ & 1 \\
        $w_e$ & Weight for $r_e$ & 1 \\
        $w_a$ & Weight for $r_a$ & 1 \\
        $dis_0$ & Maximum interaction distance of CAV & $40m$ \\
        $\delta_0$ & Attention threshold of CAV & 0.05 \\
        $\mathcal{Q}$ & Maximum number of interactive objects for a CAV & 5\\
        \bottomrule
    \end{tabular}
\end{table}

\subsection{Performance Evaluation}
\subsubsection{Overall Performance}
The average and cumulative rewards of the agent during training are shown in Fig. \ref{fig:alg_results}. In the environment just having CAVs, the performance of Attention-MADDPG and MA-GA-DDPG are both significantly better than that of MADDPG, which proves that the Attention mechanism can effectively improve the performance of the algorithm.

In the CAV-HV mixed driving environment, MA-GA-DDPG shows more obvious advantages. In the mixed driving environment of CAVs and homogeneous HVs, Attention-MADDPG has shown more powerful performance than MADDPG, while the stability and convergence of MA-GA-DDPG are further better than Attention-MADDPG. MA-GA-DDPG obtains the most cumulative rewards during training, indicating that the model agent learns a policy function that can drive safely in traffic for a longer period of time.

In the CAVs and heterogeneous HVs mixed driving environment, HVs with different driving styles will bring more complex and diverse interactive behaviors, which puts forward higher requirements for the learning ability of the model. MA-GA-DDPG still shows the best learning ability, and the average reward and cumulative reward are significantly better than the two baselines. Our proposed model exhibits significant performance improvement in these difficult scenarios.

\subsubsection{Performance in Different Scenarios}
\begin{figure*}[!htbp]
    \centering
    \includegraphics[width=1\textwidth]{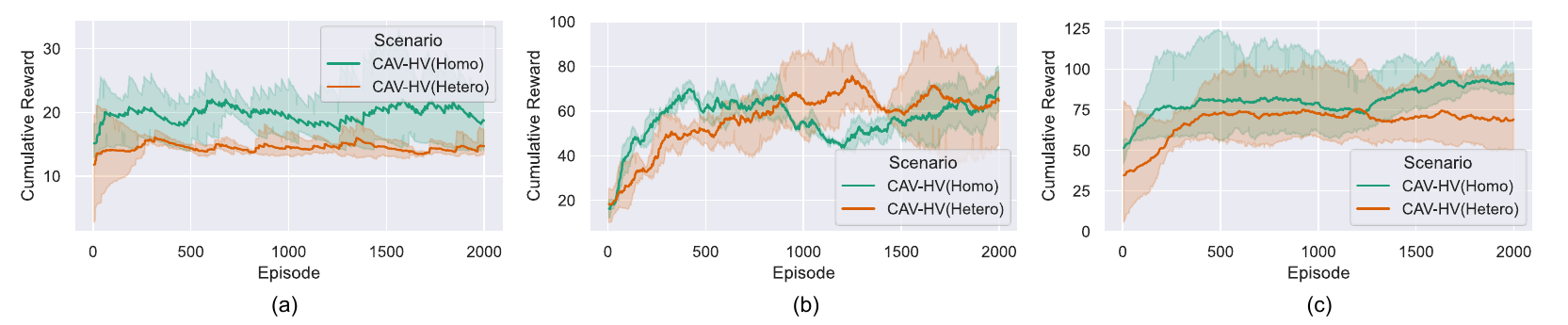}
    \caption{The influence of human driving characteristics on 
 performance of different models :(a) MADDPG, (b) Attention-MADDPG, (c) MA-GA-DDPG.
    }
    \label{fig:different_HVs_compare}
\end{figure*}
At the same time, we compared the performance of the algorithms in different mixed driving environments to analyze the impact and challenges of heterogeneous HVs on model learning. 
As shown in Fig. \ref{fig:different_HVs_compare}, in the presence of heterogeneous agents, the performance of the three algorithms varies to different degrees. Specifically, MADDPG displays a certain degree of degradation in performance, whereas Attention-MADDPG shows no significant decline. This indicates that incorporating attention-based policy networks can enhance algorithm performance in complex environments. By contrast, although the performance of MA-GA-DDPG decreases when faced with heterogeneous agents, its overall performance still surpasses that of Attention-MADDPG and MADDPG.

\subsubsection{Safety Analysis}
Safety is the most critical factor to consider when designing CAV decision-making algorithms.
We first conduct $100$ random scenario tests on three algorithms and count the success rate. When all the CAVs in the scenario do not collide and reach the destination smoothly, we record it as a success. After simulation, the success rate of MADDPG, A-MADDPG, and MA-GA-DDPG are $44.0\%$
, $72.0\%$, $86.0\%$ respectively.

\begin{figure}
    \centering
    \includegraphics[width=0.5\textwidth]{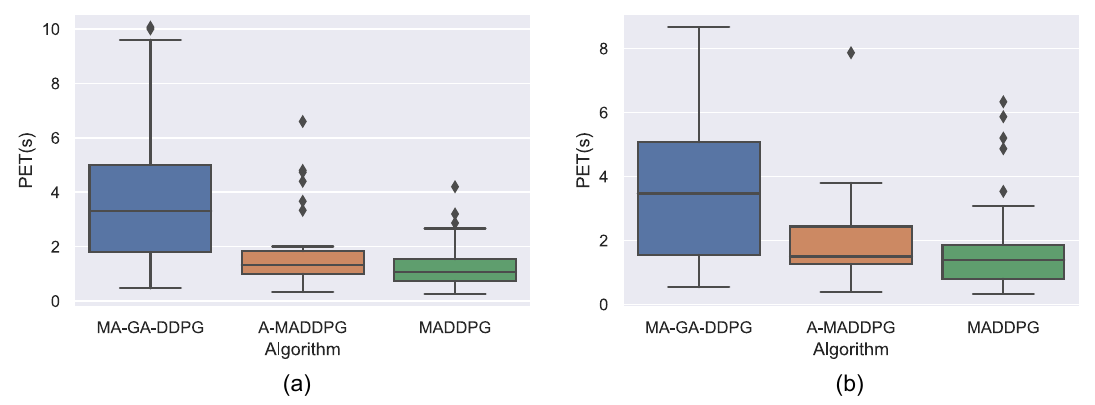}
    \caption{CAVs' PET statistics of different algorithms in different scenarios: (a)CAVs and homogeneous HVs; (b) CAVs and heterogeneous HVs.}
    \label{fig:PET_analysis}
\end{figure}

Meanwhile, to evaluate security during interactions, we employ the post-encroachment time (PET) metric\cite{ma2017two}. Fig. \ref{fig:PET_analysis} illustrates the PETs of different algorithms in various scenarios involving interactions between CAVs and other CAVs as well as all HVs. 
The average PET of MADDPG in CAH-HV (homogeneous) and CAV-HV (heterogeneous) is $1.30 s$ and $1.72 s$ respectively, and the average PET of A-MADDPG in CAH-HV (homogeneous) and CAV-HV (heterogeneous) is $1.80 s$ and $1.96 s$ respectively.
In contrast, MA-GA-DDPG yields results of $3.61 s$ and $3.59 s$, representing a $64.0\%$ and $52.1\%$ increase over MADDPG. 

Additionally, we found that different algorithms exhibit varying levels of stability when the heterogeneity of the HVs changes. Compared to the homogeneous HV environment, the average PET of MADDPG and A-MADDPG in the heterogeneous HV environment increased by $32.3\% $ and $8.9\%$, respectively, while the performance of MA-GA-DDPG was more stable, with an average PET fluctuation of only $0.5\%$.

Overall, our simulation results and analysis indicate that MADDPG is the most aggressive in the interaction process, with the poorest safety performance, while our algorithm is safer and more stable.

\subsubsection{Efficiency and Comfort Analysis}
\begin{figure}
    \centering
    \includegraphics[width=0.5\textwidth]{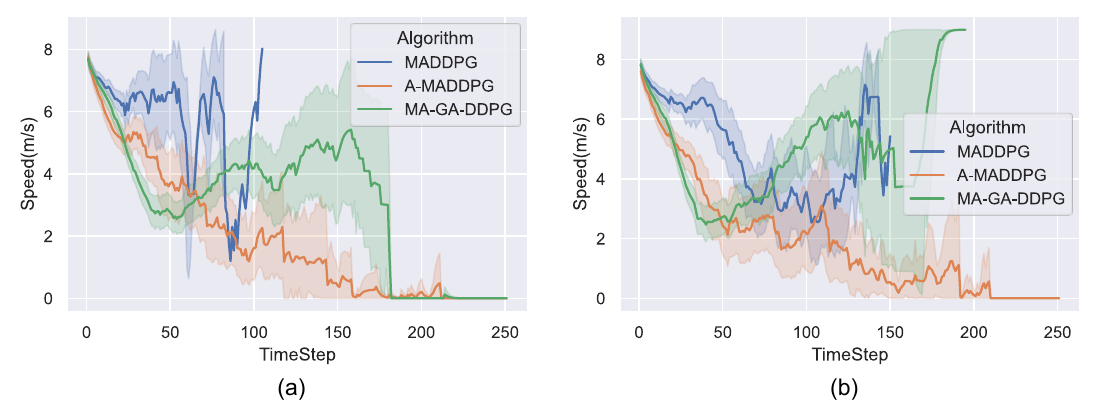}
    \caption{The average speed of CAVs from different algorithms: (a) CAVs and homogeneous HVs; (b) CAVs and heterogeneous HVs.}
    \label{fig:ave_speed}
\end{figure}

We expect that CAVs can maintain both efficiency and comfort while prioritizing safety. To gauge the effectiveness of CAVs, we examine the speed variation curve of CAVs in testing scenarios, while assessing driving comfort by analyzing longitudinal acceleration. 
As depicted in Fig. \ref{fig:ave_speed}, the MADDPG algorithm exhibits the highest average speed in both mixed driving environments; however, its aggressive driving behavior leads to dangerous interactions and frequent collisions, as previously analyzed. The A-MADDPG algorithm registers the lowest average speed, but it passes through intersections at a slow pace, resulting in decreased efficiency. On the other hand, our MA-GA-DDPG algorithm can promptly decelerate to ensure interaction safety when approaching the intersection and then accelerate appropriately after passing the conflict point, striking a balance between safety and efficiency.

\begin{figure}
    \centering
    \includegraphics[width=0.5\textwidth]{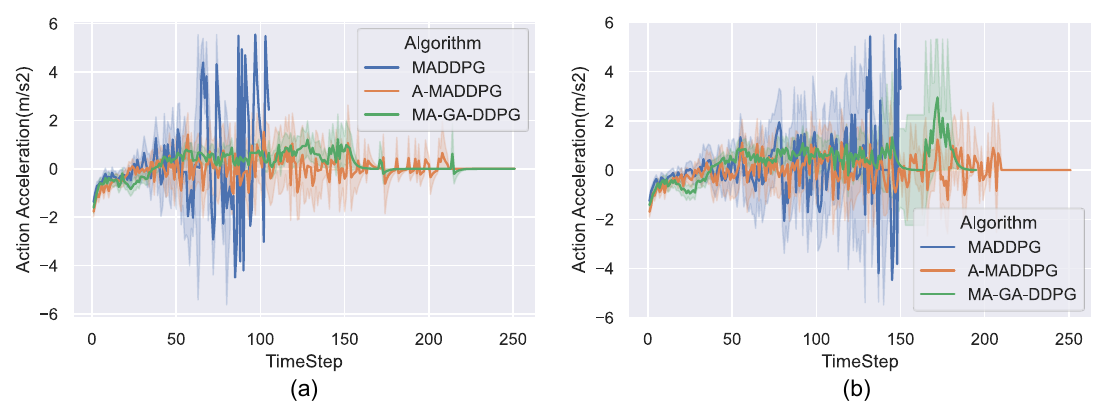}
    \caption{The average action acceleration of CAVs from different algorithms: (a) CAVs and homogeneous HVs; (b) CAVs and heterogeneous HVs.}
    \label{fig:ave_acc}
\end{figure}

The acceleration curve, shown in Fig. \ref{fig:ave_acc}, highlights that MADDPG has an excessive amplitude of acceleration and deceleration when navigating through intersections, causing significant discomfort. In contrast, the A-MADDPG and MA-GA-DDPG algorithms exhibit gentler acceleration and deceleration amplitudes, ensuring improved comfort.

\subsection{Algorithm Extension Testing and Case Analysis}
\begin{figure}[!htbp]
    \centering
    \includegraphics[width=0.5\textwidth]{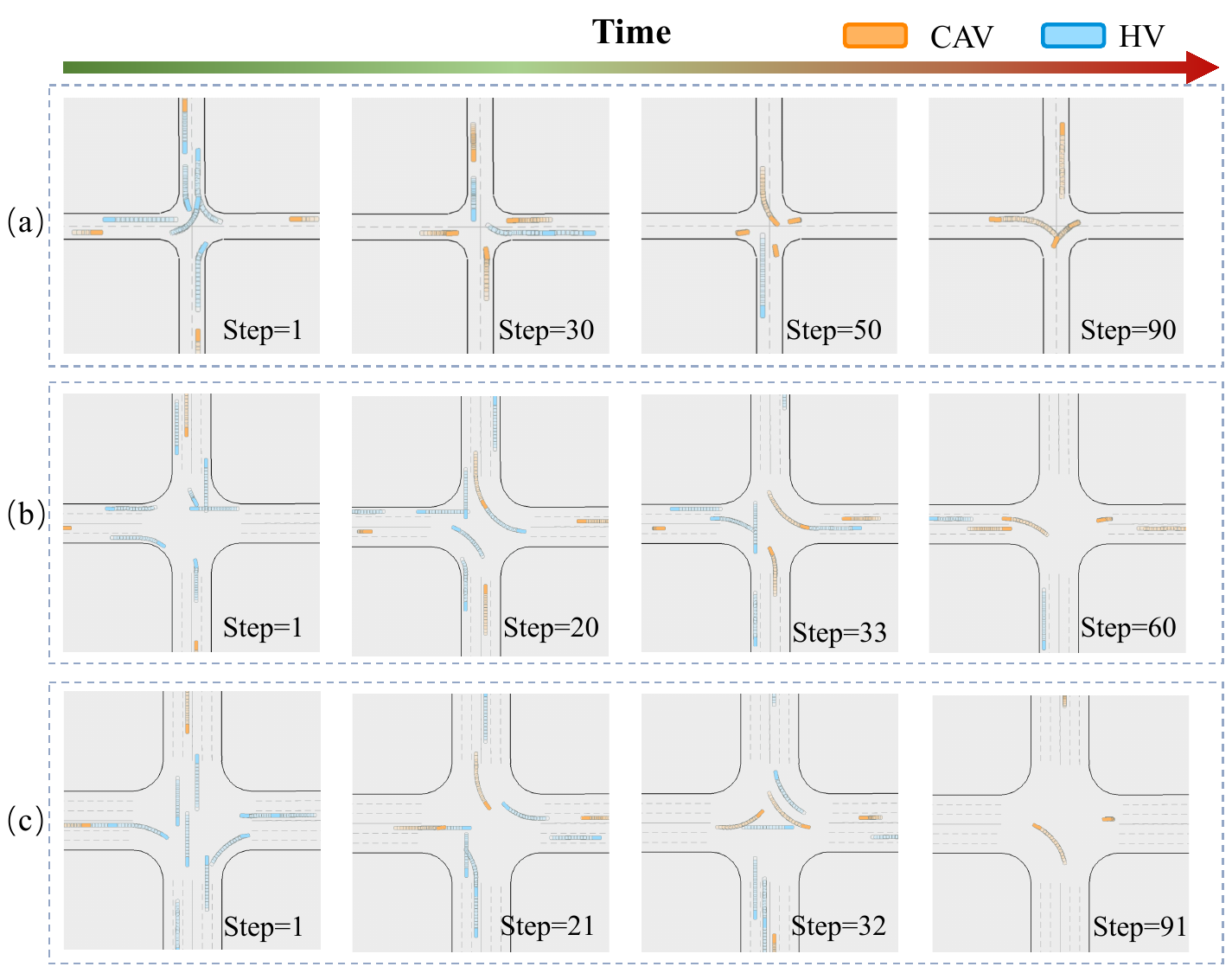}
    \caption{Snapshots at critical moments of three interaction cases, (a) case 1,single-lane intersection; (b) case 2, two-lane intersection; (c) case 3, three-lane intersection.}
    \label{fig:snapshot_interaction_process}
\end{figure}

To further test the effectiveness of our algorithm, in addition to the basic single-lane cross-shaped unsignalized intersection, we designed two more complex intersection scenarios: a two-lane unsignalized intersection and a three-lane unsignalized intersection, as shown in Fig. \ref{fig:snapshot_interaction_process} (b) and (c). These scenarios feature more complex traffic conflicts and traffic flows, providing a robust test of our algorithm.

Our experiments demonstrated that in the two-lane and three-lane intersections, our MA-GA-DDPG algorithm achieved passage success rates of 84.5\% and 82.0\%, respectively. These results highlight the effectiveness and scalability of our algorithm across different scenarios.

On the other hand, to explore the micro-behavioral interactions of CAVs, we selected interaction cases from these three scenarios for detailed analysis. Fig. \ref{fig:snapshot_interaction_process} illustrates the key moments in the interaction processes of three cases. In all cases, the CAVs from the south, west, north, and east approaches are labeled as $\text{CAV}_1$, $\text{CAV}_2$, $\text{CAV}_3$, and $\text{CAV}_4$, respectively. The animated version of three cases can be accessed at the site.\footnote{See \url{https://drive.google.com/drive/folders/1v8wtHtBeGzpuh3E-qNiGBQKEXMtn2G3K?usp=sharing}}

\textbf{Case 1} involved four CAVs and six HVs from different approaches. As the CAVs slowed down to yield, most HVs had already left the vicinity by the time the vehicles approached the intersection. At $\text{Timestep}=30$, $\text{CAV}_4$ from the south attempted a left turn but faced a potential conflict with a northbound straight-moving HV. After assessing the risk, $\text{CAV}_4$ decided to stop and yield. Following the HV’s passage, $\text{CAV}_3$ from the north made the first left turn through the intersection ($\text{Timestep}=50$). Subsequently, the remaining three CAVs negotiated and sequentially passed the conflict point at the intersection ($\text{Timestep}=90$).

\textbf{Case 2} saw $\text{CAV}_1$ at $\text{Timestep}=20$ observing an oncoming straight-moving HV and beginning to decelerate. By $\text{Timestep}=33$, $\text{CAV}_1$ stopped to wait for the HV to pass before accelerating through the intersection. At $\text{Timestep}=60$, $\text{CAV}_2$, noticing a potential conflict with $\text{CAV}_1$, also decelerated and waited for $\text{CAV}_1$ to clear the conflict point before accelerating through.

\textbf{Case 3} involved $\text{CAV}_2$ decelerating at $\text{Timestep}=21$ to allow $\text{CAV}_3$ to pass first and then accelerating after the conflict cleared at $\text{Timestep}=32$. At $\text{Timestep}=91$, $\text{CAV}_4$ also identified a potential conflict with $\text{CAV}_1$, decelerated to yield, and waited for $\text{CAV}_1$ to pass the conflict point before accelerating through the intersection, ensuring all CAVs passed safely.

Overall, the CAVs exhibited robust and cautious driving styles. Upon approaching an intersection, they would typically slow down to observe the surroundings and predict the driving behaviors and future trajectories of nearby traffic participants. Based on the conflict situation, they made various driving decisions such as accelerating, stopping to yield, or forcefully accelerating, balancing safety and efficiency while demonstrating strong interaction capabilities to smoothly navigate through the intersection.

\subsection{Hardware-in-loop Experiment}
To validate the effectiveness and practicality of our approach, we designed a hardware-in-the-loop (HIL) experiment. We incorporated the Autonomous Driving Control Unit (ADCU), also known as the domain controller, into the entire experimental setup as hardware. The ADCU replicates the computational environment of a real vehicle, which not only avoids the safety risks associated with on-road testing but also closely approximates the effects of real-vehicle tests.

\begin{figure}
    \centering
    \includegraphics[width=0.4\textwidth]{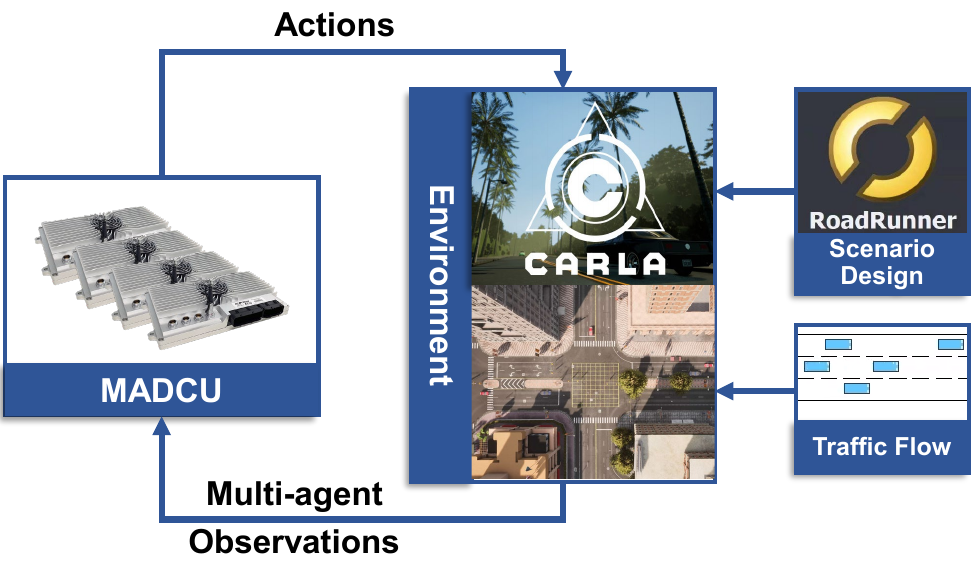}
    \caption{The framework of ADCU-in-loop experiment.}
    \label{fig:adcu_framework}
\end{figure}

\textbf{Experimental Framework}: Our domain controller-in-loop simulation framework is depicted in Fig.\ref{fig:adcu_framework}. We designed the simulation environment using the Carla simulator \cite{dosovitskiy2017carla}, configuring and injecting HVs as background traffic flow based on the highway simulator\cite{highway-env}. Several ADCUs control CAVs, with each ADCU controlling one CAV. CAVs can communicate with each other. The Carla simulator publishes all agents' perception data through the ROS system \cite{quigley2009ros}, and each ADCU subscribes to its environmental perception information as state input. The policy network processes these inputs to produce decision actions for each CAV. These actions are then transmitted back to the Carla simulator via ROS messages for action execution and the simulation of the next timestep.

\begin{figure}
    \centering
    \includegraphics[width=0.4\textwidth]{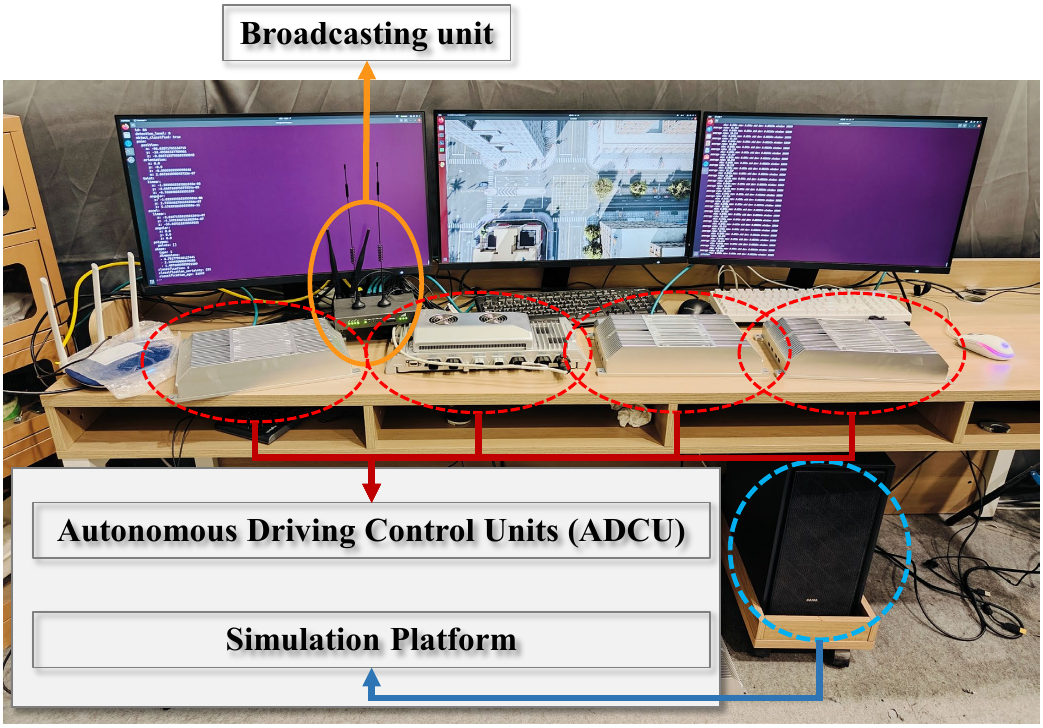}
    \caption{The field test of ADCU-in-loop experiment.}
    \label{fig:adcu_experiment}
\end{figure}

\textbf{Hardware Platform and Experimental Setup}: Our testing platform, as shown in Fig.\ref{fig:adcu_experiment}, includes four ADCUs, a computational host, broadcasting and routing units, and several monitors. We set up a two-lane unsignalized intersection environment with four CAVs, each entering the intersection from one of four directions.

\textbf{Experimental Validation Results}: In real tests, the decision algorithms deployed on the ADCUs were capable of responding and computing at a frequency of 20Hz, fully meeting the real-time computational requirements of actual vehicles. Moreover, in the hardware-in-the-loop experiments, our method achieved a passage success rate of 83.1\%. The computational accuracy and the error margin compared to pure simulation experiments were maintained within 2\%, demonstrating the efficiency and precision of our experiments. The animated version of the field experiment can be accessed at the site.\footnote{See \url{https://drive.google.com/drive/folders/150_D_ZEmzhSum_fGo-uslcm9bYz4_5Qh?usp=drive_link}}

\section{Conclusions}
\label{section:6}
Cooperative decision-making for CAVs in complex human-machine environments remains a significant challenge. This paper addresses these issues by defining a decentralized MARL problem for navigating such intersections and introduces a novel algorithm, Multi-Agent Game-Aware Deep Deterministic Policy Gradient (MA-GA-DDPG). This algorithm integrates an attention mechanism with level-k game priors, enhancing the safety and efficiency of CAVs. The attention-based policy network improves learning efficiency and captures complex interactions between the ego CAV and other agents, using attention weights as interaction priors in a hierarchical game framework, which includes a safety inspector module to enhance CAV safety. Comprehensive experiments, including simulation and ADCU-in-loop testing that consider human driver heterogeneity, demonstrate that MA-GA-DDPG significantly improves safety, efficiency, and comfort.

In the future, our research will focus on expanding our algorithm to more complex and realistic scenarios. We will also extend the prediction horizon and develop a more sophisticated conflict resolution module to enhance performance. Additionally, we will thoroughly explore the social dynamics between CAVs and human-driven vehicles, aiming to ensure that CAVs emulate human driving behaviors while prioritizing safety and efficiency.

\bibliographystyle{IEEEtran}  
\bibliography{reference}  

\vfill

\end{document}